\documentclass[10pt,twocolumn,letterpaper]{article}

\usepackage[pagenumbers]{cvpr} %

\newcommand\blankfootnote[1]{%
  \begingroup
  \renewcommand\thefootnote{}\footnote{#1}%
  \addtocounter{footnote}{-1}%
  \endgroup
}
\definecolor{cvprblue}{rgb}{0.21,0.49,0.74}
\usepackage[pagebackref,breaklinks,colorlinks,allcolors=cvprblue]{hyperref}

\DeclareMathOperator*{\argmax}{arg\,max}

\def\paperID{0000} %
\def\confName{CVPR}
\def\confYear{2025}

\title{Joint Out-of-Distribution Filtering and Data Discovery Active Learning}

\author{Sebastian Schmidt$^{1,2}$ \quad Leonard Schenk$^{3,{\dag}}$ \quad Leo Schwinn$^1$ \quad Stephan Günnemann$^{1}$\\
$^1$ Technical University of Munich, School of Computation, Information and Technology \\
 $^2$ BMW Group \quad $^3$ SPRIND
}

\begin{document}
\maketitle

\blankfootnote{$^{\dag}$ Work done while being with TUM and BMW.}
\blankfootnote{Corresponding author {\tt sebastian95.schmidt@tum.de} }

\newcommand{\ours}{Joda}
\newcommand{\topq}[2]{
  \operatorname*{arg\,max}_{x \in #1} \, #2(x)
}
\begin{abstract}
As the data demand for deep learning models increases, active learning (AL) becomes essential to strategically select samples for labeling, which maximizes data efficiency and reduces training costs.
Real-world scenarios necessitate the consideration of incomplete data knowledge within AL. Prior works address handling out-of-distribution (OOD) data, while another research direction has focused on category discovery. However, a combined analysis of real-world considerations combining AL with out-of-distribution data and category discovery remains unexplored.
To address this gap, we propose \textbf{J}oint \textbf{O}ut-of-distibution filtering and data \textbf{D}iscovery \textbf{A}ctive learning (\ours{})
\footnote{Project Page: \url{https://www.cs.cit.tum.de/daml/joda/}}
, to uniquely address both challenges simultaneously by filtering out OOD data before selecting candidates for labeling.
In contrast to previous methods, we deeply entangle the training procedure with filter and selection to construct a common feature space that aligns known and novel categories while separating OOD samples. 
Unlike previous works, \ours{} is highly efficient and completely omits auxiliary models and training access to the unlabeled pool for filtering or selection.
In extensive experiments on \textbf{18} configurations and \textbf{3} metrics, \ours{} consistently achieves the highest accuracy with the best class discovery to OOD filtering balance compared to state-of-the-art competitor approaches.
\end{abstract}

\section{Introduction}
\label{sec:Introduction}

Deep learning models, particularly in computer vision, depend on extensive labeled datasets, which entails considerable annotation costs. Active learning (AL) offers a systematic approach to reduce annotation costs by selecting only the most informative samples for labeling. AL methodology involves a cyclic process to select previously unlabeled samples utilizing auxiliary models or properties like uncertainty or diversity.

In the classic closed set AL \cite{settles2010}, an unlabeled pool or data stream is defined as \emph{pure}, containing only in-distribution (InD) data that is of the same underlying distribution as the training data.
Recent works \cite{Ning2022,Park2022,Du2023,Yang2023} propose \emph{open-set AL} to question the real-world applicability of this assumption and consider out-of-distribution (OOD) data in the unlabeled pool. In real-world applications, data is noisy and originates from different distributions. This necessitates detecting and filtering out OOD data before selecting samples for labeling.
Furthermore, dynamic and open environments contain novel objects that are absent in the training data but relevant to the task.
Discovering these new InD categories is addressed by the field of category discovery \cite{Vaze2023b}, where incomplete knowledge about the InD data is assumed. %

While existing research tends to address only one of these challenges at a time, both phenomena are likely to arise concurrently in open- or real-world applications, such as mobile robotics, environmental sensing, or autonomous driving, given the limited prior knowledge of a narrow initial dataset.
Conversely, extensive data collection in real-world scenarios will inevitably lead to encounters with OOD data and novel object categories.
Considering the data collection by an advanced driver assistance system of cars, in the endeavor to collect data for fully autonomous driving, unknown participants like a carriage reflect novel categories, while scenarios outside of the domain like off-road, private properties, or gravel roads pose OOD data.
Especially, for autonomous driving safe operation is required \cite{schmidt2024deep}.
Besides the limited open-world assumptions, existing works \cite{Park2022,Du2023,Yang2023,Safaei2024} require additional models and data access to the unlabeled pool, a challenging requirement given incomplete knowledge of its data.

\begin{figure*}[tbh]
    \vskip -0.2cm
    \centering
\includegraphics[width=\textwidth]{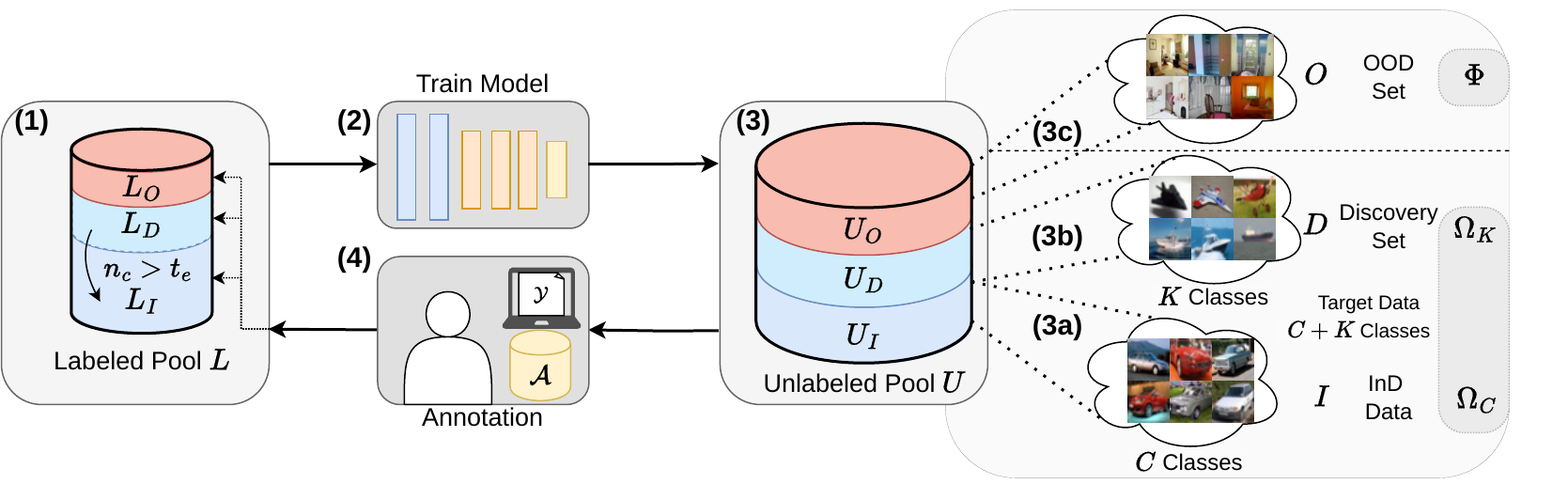}
    \vskip -0.3cm
    \caption{Overview of the Open-Set Discovery Active Learning cycle. Starting with the labeled pool \textbf{(1)} for training a model \textbf{(2)} used to select data from an unlabeled pool \textbf{(3)}. Contrasting with previous works, it comprises three subsets: known classes \textbf{(3a)}, novel discoverable classes \textbf{(3b)}, and unwanted OOD data \textbf{(3c)}. After selection, the cycles closed with annotation \textbf{(4)}.}
    \label{fig:Scenario}
    \vskip -0.3cm
\end{figure*}

Our work proposes a novel approach to jointly manage impure data assumptions of open-set scenarios \textbf{and} incomplete prior knowledge assumptions of category discovery \textbf{without} using additional models or unlabeled data.
We provide the following \textbf{contributions}: \\
\begin{itemize}
    \item We introduce \textbf{O}pen-\textbf{S}et \textbf{D}iscovery \textbf{A}ctive \text{L}earning (\textbf{OSDAL}), a novel scenario that jointly describes the impurity of the unlabeled data pool with the incomplete knowledge of the relevant classes (illustrated in \cref{fig:Scenario}).\\
    \item We propose \textbf{J}oint \textbf{O}ut-of-distribution filtering and data \textbf{D}iscovery \textbf{A}ctive Learning (\textbf{Joda}) as the first approach being able to separate OOD from novel classes that addresses these scenarios. Contrasting to existing works in open-set AL \cite{Ning2022,Park2022,Du2023,Yang2023,Safaei2024}, \ours{} \emph{eliminates} the need for auxiliary models and data access.
    \item We conduct extensive experiments on \textbf{15} settings to evaluate current AL methods and open-set AL methods. To the best of our knowledge, this is currently the largest benchmark in open-set AL. \ours{} confirms its effectiveness by consistently achieving the highest accuracy despite the considerably reduced complexity.
\end{itemize}

\section{Related Work}
\label{sec:RelatedWork}

\textbf{Active Learning:}
Generally, AL focuses on strategically selecting samples for labeling to preserve high model performance while reducing the number of required annotations in stream-based or pool-based scenarios \citep{settles2010, Zhan2022}.
Besides the classic scenarios, \citet{Schmidt2023} and \citet{Schmidt2024b} evaluated further scenarios featuring dynamic pools or batch streams.
Furthermore, the various methods employed for AL are categorized into uncertainty-based, diversity-based, or model-based. \\
Uncertainty-based methods like ensembles \cite{BeluchBcai, Lakshminarayanan2016} or Monte Carlo Dropout, which leverage multiple forward passes \cite{Gal2015, Kirsch2019} select samples based on an estimated uncertainty value.
Their individual sample evaluation favors them for various applications like 2D and 3D object detection \cite{Feng2019, Schmidt2020, Hekimoglu2022, Park2023}, semantic segmentation \cite{Huang2018} or graphs \cite{Fuchsgruber2024}. %
Diversity-based approaches aim to fully cover all regions of the dataset space \cite{Sener2018,yehuda_active_2022}. %
Badge \cite{Ash2020} combines diversity with uncertainty by gradients calculated utilizing the network's predictions as pseudo-labels.
Subsequent research follows the idea of combining uncertainty and diversity for more complex tasks like 3D object detection \cite{Yang2022, Luo2023}. %
Lastly, model-based methods employ an auxiliary model to either predict a proxy score for the sample informativeness or select samples directly. The methodology includes predicted loss values \cite{Yoo2019}, encoder decoder-based approaches \cite{Sinha2019, Zhang, Kim2021} or meta-models like graph neural networks \cite{Caramalau2021} as well as teacher-student approaches \cite{Peng2021, Hekimoglu2024}. In this work, we go beyond the standard methodology used in AL and introduce aspects of OOD detection in our \ours{} algorithm.

\textbf{Out-of-Distribution Detection:}
The term OOD detection is a generic term frequently used to encompass a range of related concepts, including anomaly detection, novelty detection, and open-set recognition (OSR) \cite{Yang2021,salehi_unified_2021}.
In a common understanding of OOD detection, a sample must be categorized into InD or OOD.
Approaches typically employ uncertainty-based methods or feature space methods. Uncertainty-based methods include ensembles \cite{arpit_ensemble_2022}, energy scores \cite{liu_energy-based_2020, elflein2021outofdistribution,Fuchsgruber2024b}, density estimation \cite{Charpentier2020, Charpentier2022}, and entropy \cite{x_liu_y_lochman_c_zach_gen_2023}. Feature space methods involve latent space distances \cite{Sun2022,lee_simple_2018} or gradient-based methods \cite{liang_enhancing_2020, Hsu2020, huang_importance_2021, pmlr-v161-schwinn21a, liu_neuron_2023}. The dual SISOMe approach of \citet{Schmidt2024} combines uncertainty and feature space metrics, making it robust across different settings. \ours{} uses SISOMe for sample selection and extends it with a task adaptive training algorithm, an OOD filtering step, and a score re-balancing.

\textbf{Open-Set Active Learning:}
Open-set AL extends the classic AL scenario by incorporating OOD samples in the unlabeled pool. The methodology primarily relies on methods that treat sample selection and OOD filtering separately, requiring one model for classifying InD samples and an additional model for identifying OOD data \cite{Ning2022,Du2021,Du2023,Park2022,Yang2023,Safaei2024,Zong2024}. For example, \citet{Ning2022} used a Gaussian Mixture Model for the AL sample selection and an auxiliary classification model to detect and filter out OOD samples. \ours{} completely refrains from additional models or data.

\textbf{Category Discovery:}
Category discovery assumes an incomplete knowledge of the existing categories in the dataset \cite{Cao2022}. All classes are considered to be within the same domain, but not all classes are known a priori. Within this task, approaches share various techniques with OOD detection, such as contrastive learning \cite{Vaze2023b, Fei2022,Wen2023}. To unveil new categories, latent space clustering approaches like k-means \cite{Vaze2023b,Chiaroni2023} or teacher-student approaches \cite{Vaze2023a} are commonly used.
\citet{Ma2024} introduced the task of Active Category Discovery (AGD), a combination of category discovery and AL, which exclusively queries novel categories.

\textbf{Concluding}, existing works addressing AL and OOD detection do not cover category discovery.
Conversely, \citet{Ma2024} addressed category discovery but only for novel classes, not for known or OOD data. The combination of AL, OOD detection, and category discovery remains unexplored and will be addressed in this work.

\section{Open-Set Discovery Active Learning}
\label{sec:Scenario}
This section introduces the novel Open-Set Discovery Active Learning (OSDAL) scenario, combining AL, OOD data, and category discovery. We illustrate this novel scenario in \cref{fig:Scenario}.
We begin with a labeled pool $L$ \textbf{(1)} and an unlabeled pool $U$ \textbf{(3)}. In each iteration $i$ of the cyclic AL process, a set $\mathcal{A}$ of query size $q$ is chosen from $U$ and sent to an annotator \textbf{(4)}. The annotated set $\mathcal{A}$ is then added with its corresponding labels to $L$ for the subsequent cycle, such that $U^{i+1}= U^i \setminus \mathcal{A}$ and $L^{i+1}= L^i \cup \mathcal{A}$. In each cycle, the model \textbf{(2)} $f$ is (re-)trained such that it minimizes the loss over the labeled data, i.e., $\displaystyle f(\omega): \min_{\omega} \mathcal{L}(x, y) $, where \((x, y) \in (\mathcal{X}_L, \mathcal{Y}_L)\), and \(\mathcal{X}_L\) represents the set of labeled samples while \(\mathcal{Y}_L\) represents their corresponding labels. This model enables the query strategy $Q(\mathcal{X},f)$ to choose data based on the model output or intermediate representation $f$. We extend the classic AL settings by assuming incomplete relevant class knowledge and OOD data pollution. To do so, we define three sets of data types: the InD data $I$ containing the initial known categories \textbf{(3a)}, the discovery set $D$ retaining the novel discoverable categories \textbf{(3b)}, and the OOD set $O$ \textbf{(3c)} holding data that belong neither to known nor unknown relevant categories.

To establish the discovery set $D$ and the OOD set $O$, we orientate on the distinction between near- and far-OOD data commonly employed in OOD benchmarks \cite{yang2022openood,zhang2023openood}. Near-OOD data is semantically more similar to InD data, making differentiation more challenging. On the other hand, far-OOD data differs more strongly from InD data, making it easier to distinguish. In the presented scenario, the categories to be discovered are assumed to be related to near-OOD data, while far-OOD data relates to the OOD data. We can consider $I$ and $D$ to be sampled from true category distribution $\Omega$, while only the initial distribution $\Omega_C$ containing $C$ categories has been observed given the samples in $L$. The remaining $K$ categories can be sampled from $\Omega_K$ as samples of $D$ and should be discovered throughout the cyclic process, with the number of discoverable classes $K$ remaining unknown. In contrast, we assume far-OOD data to be sampled from a distribution $\Phi$ containing irrelevant classes for the application that are neither categorized as any of the $K+C$, which should be filtered out, akin to open-set AL.

As this subset categorization of $U$ is unknown, a query strategy must identify the informative samples in $U$ to unveil the unknown categories and improve the model performance on $C+K$ categories. In each cycle, a set $\mathcal{A}$ of $q$ samples can be selected and sent for annotation \textbf{(4)}.
Subsequently, these samples are added to the labeled pool $L$, which is divided into three sub-pools: $L_O$, containing the selected OOD data identified by the annotator; $L_I$, comprising samples of already known classes; and $L_D$, designated for newly discovered classes where the number of samples $n_c$ is below a threshold $t_e$.
If the sample count of a category reaches the threshold $t_e$, it will be included in the known classes and transferred to $L_I$.
This threshold simulates a triage process by the annotator to justify the relevance and prevents training instabilities due to a highly unbalanced training set. Alternatively, the threshold can be set to zero, and techniques for strongly unbalanced datasets, such as uniform sampling classes, can be applied.
Aligned with the open-set AL scenarios \cite{Ning2022}, the (accidental) queried OOD data $L_O$ marked by the annotator is available for future use in the next cycle.
The extension of the labeled pool $L$ \textbf{(1)} with the queried data $\mathcal{A}$ initiates a new cycle, and the model is retrained.
\begin{figure*}[tbh]
    \centering
    \includegraphics[width=\linewidth]{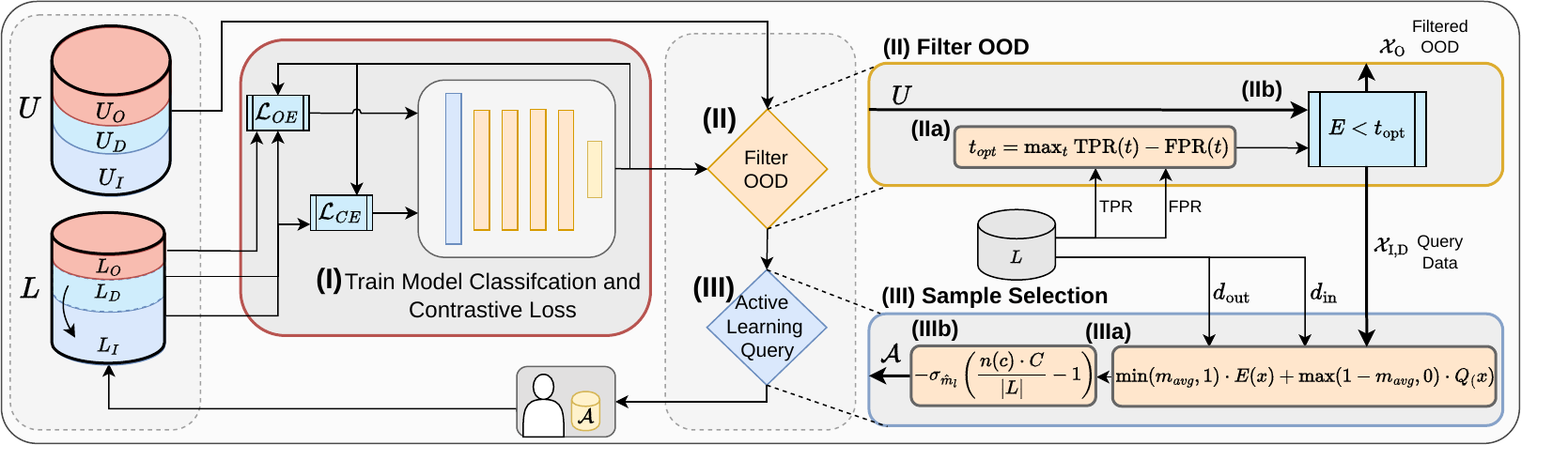}
    \vskip -0.3cm
    \caption{Joint Out-of-Distribution Filtering and Data Discovery Active Learning, comprising of the training phase \textbf{(I)} combining classification and outlier exposures loss followed by the filtering \textbf{(II)} and selection phase \textbf{(III)}. For the filtering, a threshold is estimated on $L$ \textbf{(IIa)} to separate OOD samples based on their energy value \textbf{(IIb)}. Subsequently, samples are selected based on the SISOMe \cite{Schmidt2024} metrics \textbf{(IIIa)} combined with a class balancing \textbf{(IIIb)}.}
    \label{fig:method}
    \vskip -0.2cm
\end{figure*}
\section{Joint Out-of-Distribution Filtering and Data Discovery Active Learning}
\label{sec:Joda}
\begin{table}[bth]
\centering
\caption{Overview of existing open-set AL methods, with number of classes K, labeled set L and unlabeled pool U. Additional hyperparameters are separated between metric and models if possible and estimated from the official implementations. }
\vskip -0.1cm
\begin{tabular}{lccr}
\toprule
Method & Additional & Data & Additional \\
& Models & Usage & Hyperparameters \\
\midrule
LfOSA \cite{Ning2022} & 1 & \textbf{L} & 4 + 2 \\
CCAL \cite{Du2023}  & 2 & L \& U & 2 + 14 \\
MqNet \cite{Park2022}  & 2 (+1) & L \& U & 1 + ~ 14 \\
Pal \cite{Yang2023}  & 2 & L \& U & 3 + 9 \\
EOAL \cite{Safaei2024} & 2(+K) & L \& U & 8 \\
\textbf{Joda (ours)} & \textbf{0} & \textbf{L} & \textbf{1 + 0} \\
\bottomrule
\end{tabular}
\label{tab:method_comp}
\end{table}

The novel OSDAL scenario requires distinguishing between known $I$, task-relevant discoverable $D$ samples, and task-irrelevant OOD samples $O$. As we will show in various experiments, previous works \cite{Ning2022, Du2023, Park2022,Safaei2024} are not fine granular enough for the simultaneous separation of three sets. In addition, existing methods employ one or more auxiliary models that need to be optimized and mostly require access to the unlabeled pool $U$. \cref{tab:method_comp} compares the additional model and data requirements. Optimizing multiple models adds complexity during the training process, especially when $U$ with unknown composition and OOD data is used for training. In addition, it poses a huge limitation in usability for stream-based AL scenarios \cite{settles2010,Schmidt2023}.

We propose \ours{} to provide a more granular selection strategy for separating $I$, $D$, and $O$ simultaneously while eliminating the need for additional models and data.
Instead \ours{} simplifying compared to existing works and addresses the challenges of OSDAL by utilizing a \emph{single model} that is composed of well-coordinated and interdependent components.
In \cref{fig:method}, \ours{} is outlined, comprising of a training \textbf{(I)}, filtering \textbf{(II)}, and selection \textbf{(III)} phase. During the training of the task model \textbf{(I)}, \ours{} utilizes InD and unintentionally selected OOD data to separate both distributions' feature space representation, improving the subsequent OOD filtering \textbf{(II)}. As mentioned before, no auxiliary model is trained during this process. After the model is trained, we conduct an OOD filtering \textbf{(II)} leveraging our training scheme and sample selection \textbf{(III)} to identify samples for labeling.

\textbf{Training Phase (I):}
Prior research in the OOD detection domain only considers the binary separation between InD and OOD. It has not accounted for the threefold separation of InD, near-OOD, and far-OOD data.
The challenge in this scenario is to differentiate the unknown sets $D$ and $O$ to avoid mistakenly selecting OOD data or inadvertently dismissing discoverable classes.
Since the separation is not infallible, three distinct sets, $L_I$, $L_D$, and $L_O$, may appear in the labeled pool $L$.
Utilizing the existence of $O$ in the labeled pool $L$, we aim to construct a feature space that groups samples from $I$ and $D$ closer together than samples from $O$. \\
To achieve this, we propose a new loss function operation in a polluted labeled pool. We combine an outlier exposure loss \cite{hendrycks_baseline_2018} $\mathcal{L}_{OE}$ for OOD data $L_O$ within our labeled pool with a Cross-Entropy loss $\mathcal{L}_{CE}$ for the InD part  $L_I$ weighted by a hyperparameter $\lambda_{\text{OE}}$.
For each batch, we process the InD part $b_{InD} \subset L_I$ and the OOD part $b_{OOD} \subset O_L$ separately, such that the final loss function can be written as:
\begin{equation}
\begin{aligned}
    \mathcal{L}(b) &= \mathcal{L}_{\text{CE}}(b_{\text{InD}}) + \lambda_{\text{OE}} \cdot \mathcal{L}_{\text{OE}}(b_{\text{OOD}}) \\ \mathcal{L}_{\text{OE}}(b) &= - \frac{1}{b} \sum_{\substack{x \in b}} \left( \frac{1}{C} \sum_{\substack{i = c}}^{C-1} f(x)_i - \log\left(\sum_{\substack{i = c}}^{C-1}  e^{f(x)_i}\right) \right )
\end{aligned}
    \label{eq:loss}
\end{equation}
With $\mathcal{L}_{\text{OE}}$, we can regularize the model to predict a uniform distribution for OOD instances, which we will exploit in the following phases.
Given a pure labeled set $L$ containing only InD data $I$, the loss collapses to a standard cross-entropy loss.

\textbf{Separation Phase (II):}
After completing the training phase, we proceed to the querying phase with the primary objective of filtering out any OOD data before selecting data for labeling.
We utilize the energy score, as defined in \cref{eq:Energy}, as a metric for identifying the OOD samples. This approach leverages from the $\log\sum \exp$ term in \cref{eq:loss}.
\begin{equation}
E(x) = -\log\sum_{i=1}^{c} \exp(f(x)_i)
\label{eq:Energy}
\end{equation}
To determine the optimal threshold for identifying OOD data from $O$, we conduct a Receiver Operating Characteristic (ROC) analysis on the labeled pool $L$. This analysis helps to calculate the true- and false-positive rates (TPR/FPR) across various thresholds. We then identify the threshold $t_{opt}$ that maximizes Youden's J statistic \cite{Youden}, which is the difference between TPR and FPR:
$t_{opt} = \argmax \text{TPR}(t) - \text{FPR}(t)$.
This optimization identifies the top-left-most point on the ROC curve, representing the best trade-off between capturing OOD outliers and preserving InD data. Consequently, we classify unlabeled samples $x$ as InD or discoverable if their energy scores are below the threshold $\mathcal{X}_{\text{I,D}} = \{ x \in U \mid E < t_{opt} \}$ and as OOD otherwise $\mathcal{X}_{\text{OOD}} = \{ x \in U \mid E > t_{opt} \}$. Given that the initial set $L$ in the first cycle does not contain OOD data, we omit the filtering step and start with the sample selection directly.

\textbf{Selection Phase (III):}
After separating the OOD data, we proceed with selecting $q$ samples for labeling.
During the selection, it is crucial to balance the exploration of new categories with the selection of valuable samples from known classes.
To accomplish this, we employ SISOMe \cite{Schmidt2024} as a selection metric, which harnesses the ambiguity of near-OOD and valuable AL samples.
It is, therefore, suitable for selecting known and novel categories within the unlabeled pool $U$.
The SISOMe score $\hat{m}(x)$ balances the quotient $Q$ of inner and outer class distances $d_{in}$ and $d_{out}$, and energy score $E$ through a feature space separability value $m_{avg}$:
\begin{equation}
\begin{aligned}
&\hat{m}(x) = \min(m_{avg}, 1) \cdot E(x) + \max(1 - m_{avg}, 0) \cdot Q(x) \\
&Q(x) = \frac{d_{in}(x)}{d_{out}(x)};\quad m_{avg} = \frac{1}{|L|} \sum_{x} \frac{d_{in}(x)}{d_{out}(x)}; \quad x \in \mathcal{X}_{\text{I,D}}.
\end{aligned}
\label{eq:sisomE}
\end{equation}
The self-balancing through $m_{avg}$ enables the SISOMe score to reflect a diversity, uncertainty trade-off.
By evaluating the energy as well as the feature space in $\hat{m}$, we close the loop to our loss function in \cref{eq:loss}.
Since $D$ is not included in the objective of \cref{eq:loss}, it is not regularized, which promotes the exploration of novel classes.
If a class originates from the initial known classes, it accumulates more samples in subsequent cycles compared to a newly revealed class.
To address the substantial class imbalance that arises when a new class is introduced, we establish a class-balancing factor \( b_f(c) \).
By this factor, we aim to weigh the class scores to favor uniform class partitions.
We estimate the quotient of the counts per class \( n(c) \) by using the currently predicted pseudo-class \( \argmax(f(x)) \). We multiply this by the number of classes \( C \) and divide it by the size of the labeled pool \( |L| \). By subtracting $1$, we get a positive score if the number of classes is below their uniform class share $\frac{|L|}{C}$ and a negative if it is above. By scaling with the standard deviation of \( \hat{m}_l \) we create the class balance corrected score $\hat{m}_{b}(x)$:
\begin{equation}
\begin{aligned}
\hat{m}_{b}(x) = \hat{m}(x) + b_f(\argmax_c(f(x))) \\ b_f(c) = - \sigma_{\hat{m}_l} \left( \frac{n(c) \cdot C}{|L|} - 1 \right)
\end{aligned}
\label{eq:balanceFactor}
\end{equation}

Finally, we select the samples with the highest score $\topq{\mathcal{\mathcal{X}_{\text{I,D}}}}{\hat{m}_{b}}$ according to the query size $q$.
These samples are then sent for annotation which completes the AL cycle.

\section{Experiments}
\label{sec:Experiments}

Our experiments aim to measure the performance of existing AL approaches and \ours{} on the proposed OSDAL scenario. To this end, we employ the three standard benchmark sets for open-set AL, namely CIFAR-10, CIFAR-100 \cite{cifar}, and TinyImageNet \cite{Deng2009,le2015tiny}. In the classic open-set AL setting, the original datasets would be divided into one InD and one OOD dataset. In contrast, for the OSDAL setting, we partition the original dataset into InD $I$ and discovery part $D$. %
Moreover, we incorporate an additional dataset to represent the OOD data $O$. Inspired by \cite{yang2022openood}, we use randomly generated noise samples, MNIST \cite{lecun_gradient-based_1998}, Places365 \cite{zhou2017places}, and ImageNetC-800 \cite{Hendrycks2019} with the remaining classes absent in TinyImageNet as OOD data. For Places365, we use the version of OpenOOD \cite{yang2022openood} which is free of semantic overlaps. Details about the datasets are provided in \cref{app:experimentdDetails}.

\textbf{Setup:}
For the experiment settings, we follow the classic AL choices \cite{Yoo2019} and employ the predominantly used ResNet18 \cite{He2016} model.
To evaluate the InD and discovery performance, we monitor the overall accuracy on target classes, assuming an initial accuracy of 0 for any undiscovered class.
This metric effectively gauges the overall performance of the model reflecting class discovery.
Furthermore, we calculate the precision score to assess the OOD filtering quality of different algorithms. In the precision calculation, $I$ and $D$ count as positive selection while $O$ counts as negative selection.
Finally, we evaluate the number of identified classes during training to measure the discovery capabilities of the methods.
Further experiments and additional details regarding experiment setting and hardware, are presented in \cref{app:Study} and \cref{app:experimentdDetails}.%

\begin{figure*}
    \centering
    \includegraphics[width=1\linewidth]{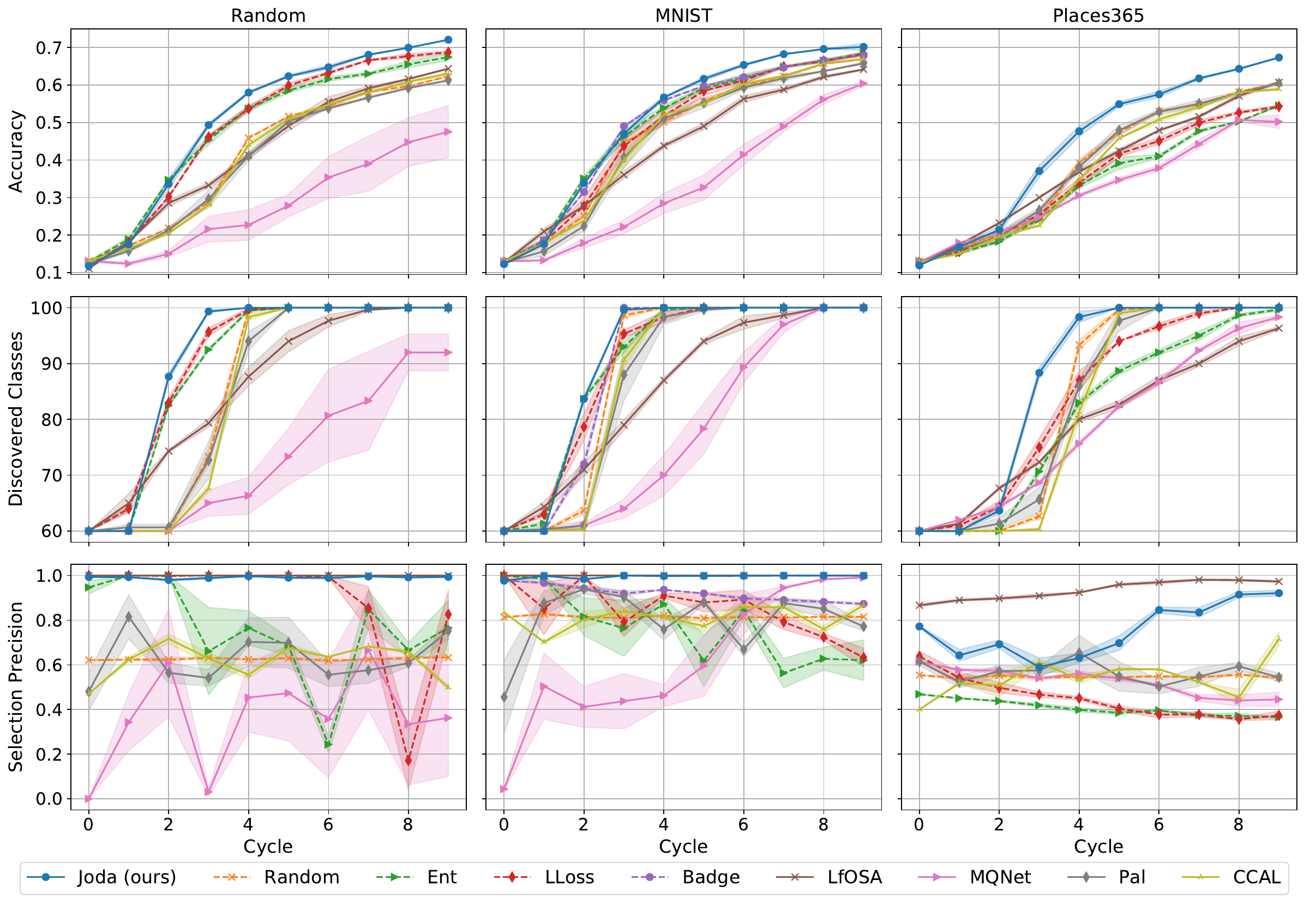}
    \vskip -0.2cm
    \caption{Comparison for CIFAR-100 with ResNet18 and indicated standard errors. From top to bottom: Mean Accuracy, Class Detection, and Selection Precision. OOD datasets from left to right: Random, MNIST, and Places365}
    \label{fig:Cifar100}
    \vskip -0.3cm
\end{figure*}

For the comparison, we encompass the following open-set AL baselines: \textbf{MQNet} \cite{Park2022}, \textbf{CCAL} \cite{Du2023}, \textbf{Pal} \cite{Yang2023}, and \textbf{LfOSA} \cite{Ning2022}, as well as the classic AL methods, Monte Carlo dropout with entropy (\textbf{Ent}) \cite{Gal2015}, Loss Learning (\textbf{LLoss}) \cite{Yoo2019}, \textbf{Badge} \cite{Ash2020}, and \textbf{Random}. \textbf{EOAL} \cite{Safaei2024} and \textbf{AGD} \cite{Ma2024} are not applicable to for OSDAL which we discuss in \cref{app:experimentdDetails}.
The open-set AL baselines consider only data from $O$ as OOD  in their auxiliary model training to avoid bias against $D$.
To enable a fair comparison, we add OOD samples from $O$ in the initial labeled pool $L$, which is required by MQNet and CCAL, although \ours{} and LfOSA can start without them.

\textbf{CIFAR-100:}
In \cref{fig:Cifar100}, we present our results for CIFAR-100. The columns represent the OOD datasets from left to right: random noise (Random), MNIST, and Places365. The visualizations depict the accuracy, class discovery, and selection precision from top to bottom.
The classic AL methods, Ent and LLoss, demonstrate decent performance for random noise as OOD but decrease in performance with the complexity of the OOD data set and get outperformed by others for Places365. Badge, which incorporates diversity, achieves decent overall performance in OOD datasets. Given that the cubic computing complex depends on the number of classes and samples, Badge is not reported for Places365 and random noise due to memory issues.
Additionally, classic AL methods produce inconsistent results for precision selection. However, together with Random selection, they benefit from a small OOD ratio, as in the case of MNIST.
Notably, the open-set methods based on contrastive learning CCAL and MQNet achieve inconsistent results. CCAL performs well for MNIST and Places365, while MQNet has difficulties with OOD Random and MNIST. For Places365, MQNet achieves good performance and class separation but suffers in selection precision.
LfOSA, utilizing an additional classifier, demonstrates good precision but struggles with class discovery.
However, Pal, CCAL, and MQNet improve their selection precision but do not surpass LfOSA or \ours{} in this aspect.

For all datasets, \ours{} exhibits a strong performance and
consistently outperforms all other methods. In the first one or two cycles, \ours{} needs a warmup. In all other cycles, \ours{} shows an early class discovery with stable selection precision. For the easier OOD data, random noise and MNIST, \ours{} show an almost perfect section precision of 1, while detecting novel categories as fastest method.
\begin{figure*}
    \centering
    \includegraphics[width=1\linewidth]{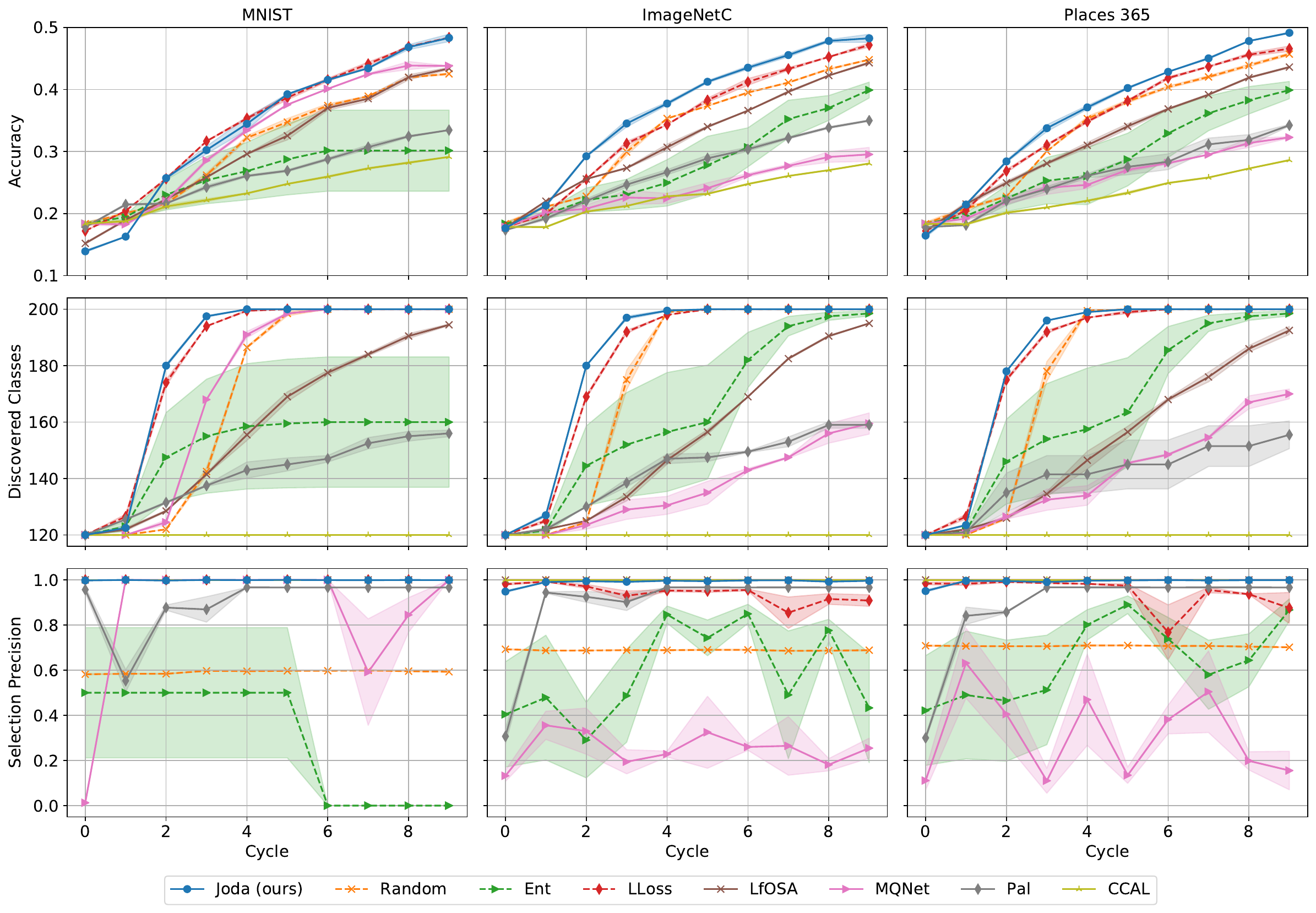}
    \vskip -0.2cm
    \caption{Comparison for TinyImageNet with ResNet18 and indicated standard errors. From top to bottom: Accuracy, Class Detection, and Selection Precision. OOD datasets from left to right: MNIST, ImageNetC-800, and Places365. }
    \vskip -0.3cm
    \label{fig:ImageNet}
\end{figure*}

\textbf{TinyImageNet:}
In \cref{fig:ImageNet}, we evaluate our scenario on the more complex TinyImageNet dataset. These experiments present the most difficult setting. The ImageNetC-800 dataset has no semantic overlap and covariate shift to TinyImageNet. The Places365 contains a minor semantic overlap with a different image style as a covariate shift, showcasing the boundaries of the ODSAL scenario. LLoss and Ent perform well for the rather artificial OOD data MNIST. While LLoss can maintain performance for ImageNetC-800 and Places365, Ent falls behind. MQNet manages high precision and selection accuracy for MNIST but drops in more realistic settings. The increased amount of samples and classes leads to issues when computing Badge. LfOSA, Pal, and CCAL achieve high selection precision but struggle with class discovery. Specifically, CCAL has issues with unveiling classes.
Remarkably, \ours{} achieves an almost perfect selection precision of $1.0$ in every cycle over all OOD datasets. Accompanied by this \ours{} shows the fastest class discovery. For the higher resolution ImageNet dataset, \ours{} gets in a competition for the easily distinguishable dataset MNIST. For the more realistic ImageNetC-800 \ours{} reaches the highest accuracy. Even for the minor covariate shift of Places365 \ours{} maintains its strong performance in accuracy, class discovery, and selection precision.
In \cref{ap:OverlapScenario} we report an additional experiment with random noise as OOD data. Given the higher image resolution, the experiment shows behavior similar to that of MNIST.

\textbf{CIFAR-10:}
Lastly, we perform experiments on CIFAR-10, the most commonly used AL benchmark. Compared to CIFAR-100 and TinyImageNet, it has a much lower number of classes. In \cref{fig:Cifar10} in \cref{ap:cifar10}, we conduct the same experiment setup as for CIFAR-100. Compared to the CIFAR-100, the margins between the methods decrease, creating competition in the first cycles. Lastly, \ours{} can achieve the highest accuracy over all OOD data sets. Again, it achieves for random noise and MNIST a selection precision of almost 1 while discovering all classes as first.

\textbf{Additional Models:}
In \cref{ap:AdditionalModelEvaluation}, we evaluate the transferability of \ours{} to a different model. For ResNet50, \ours{} can obtain a similar performance as delivered in \cref{fig:Cifar100}, enabling \ours{} to achieve the highest performance.

\textbf{Scenario Variation:}
Besides the complexity of the dataset, the number and proportion of known and discoverable categories significantly impact method behavior.
In \cref{ap:ScenarioVariation}, we evaluate the different InD to OOD ratios, different InD to novel categories ratios as well as an imbalance in number of samples per class.
 We analyze varying amounts of initially known categories in \cref{fig:CompInDRate}, where \ours{} shows a consistent performance over different rates. In experiments varying the InD to OOD setting in \cref{fig:CompOODRate}, \ours{} can continue its high performance, such that \ours{} shows a robust behavior for different InD, novel categories, and OOD ratios. In the unbalanced setting in \cref{fig:UnbalancedCounts}, \ours{} shows is robustness against other methods in this setting and increases the gap to other methods compared to \cref{fig:Cifar100}. In addition, in this setting the effect of the balancing factor introduced in \cref{eq:balanceFactor} reviles it full advantage and is analyzed.

\begin{figure*}
    \centering
    \includegraphics[width=1\linewidth]{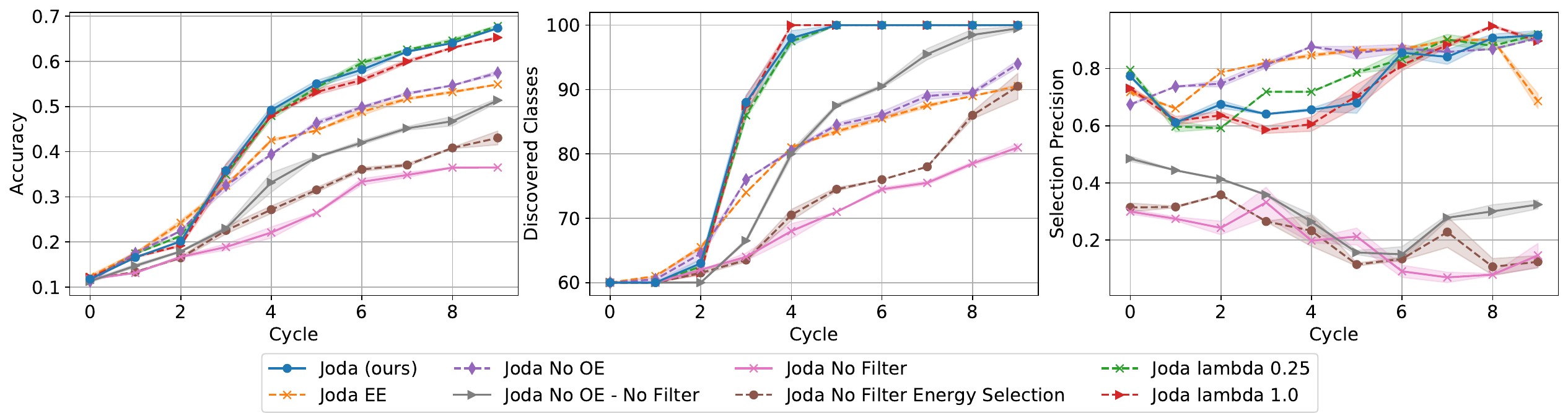}
    \vskip -0.2cm
    \caption{Ablation study on \ours{} using ResNet18 and CIFAR-100 and Places365 with indicated standard errors.}
    \vskip -0.2cm
    \label{fig:ComponentAnalysis}
\end{figure*}

\textbf{Different Query Sizes:}
The effect of different query sizes is analyzed in \cref{ap:SelSize} where \ours{} remains performant over different selection size variations.

\textbf{Ablation Study:}
In our ablation study, we investigate the different components of \ours{}, as well as different hyperparameter settings.
In \cref{fig:ComponentAnalysis}, we omitted the OE component in \cref{eq:loss}, replaced it with an Energy Exposure (EE) \cite{liu_energy-based_2020} component, and removed the energy-based filtering step (No OE - No Filter). The OE and the critical components of filtering heavily affect class discovery and selection precision and accuracy. The direct energy regularization by an EE loss component performs worse than the OE. Removing the OOD filtering and replacing the SISOMe selection with a pure energy-based selection show significant drops in accuracy, class discovery, and selection precision.
This underlines the deep entanglement of all components of \ours{} leading to a strong performance.
In addition, we compare with different values for the hyperparameter $\lambda$ in \cref{eq:loss} with the value of $0.5$ chosen for Joda. It can be seen that \ours{} shows, in general, a robust behavior against hyperparameter changes, and different variations of $\lambda$ have only a minor impact on the performance of \ours{}.

\textit{Over all experiments}, we observed a suitable performance of uncertainty-based classic AL methods for CIFAR-10 and CIFAR-100, especially LLoss, which reported a strong performance for easy OOD data and also for TinyImageNet. For Ent, we assume that smaller variation for far-OOD data leads to good filtering properties. Driven by the auxiliary open-set classifier, LfOSA consistently achieved high selection precision, which comes at the cost of a reduced class discovery. The open-set methods CCAL and MQNet employ multiple auxiliary models and achieve an inconsistent performance for the different dataset combinations, underlining the sensitivity of contrastively trained auxiliary models. Pal manages to maintain a good trade-off between exploration and selection precision.  CCAL, MQNet, and Pal struggle to distinguish OOD data from InD data if it comes from a different dataset source. \\
In 8 out of 10 combinations, \ours{} achieves a selection precision of almost $1.0$ while providing the fastest class discovery in all settings. In the 2 remaining settings, the only method showing a higher selection precision cannot unveil all unknown classes. Additionally, for 14 combinations, \ours{} achieves the highest accuracy and only ranks second for TinyImageNet with Random noise at a performance similar to the best competitor. It should be noted that especially for the most complex and realistic OOD datasets, Places365 and ImageNetC-800, \ours{} consistently reports the best accuracy over all three InD datasets.
In variation scenarios, examining different InD to OOD and InD to novel class ratios, \ours{} shows a consistent performance regardless of the ratio. The consistent performance of \ours{}, when different hyperparameters are applied, makes it easily transferable to more scenarios.

\textbf{Limitations}: The generality of our novel problem framework, OSDAL, allows for a vast range of benchmarking opportunities. While we believe that we present the most relevant evaluation, we acknowledge that future research may expand upon our work through additional datasets and OOD scenarios. \\

\section{Conclusion}
\label{sec:Conclusion}
To address AL in real-world applications like autonomous driving or environmental perception, where prior knowledge of the unlabeled data is absent, we propose \textit{Open-Set Discovery Active Learning} (OSDAL). OSDAL combines AL with the open-world challenges of OOD detection and category discovery, providing an extension of open-set AL.
We overcome these challenges by presenting \textbf{J}oint \textbf{O}ut-of-distribution filtering and data \textbf{D}iscovery \textbf{A}ctive Learning (Joda). Joda is the first AL approach with a selection capable of simultaneously filtering OOD and class discovery. Additionally, \ours{} eliminates the need for additional models required by existing open-set AL approaches. Furthermore, \ours{} only uses data from the labeled pool for training.
These properties make \ours{} lightweight and easy to apply.
In an extensive evaluation comprising \textbf{15} scenarios  \textbf{3} metrics and \textbf{8} competitors over a wide range of datasets, \ours{} consistently achieved the highest scores.

\textbf{In future work}, we aim to explore further settings and extend OSDAL and \ours{} to more complex tasks with complex novel class behavior like semantic segmentation or object detection.

{
    \small
    \bibliographystyle{ieeenat_fullname}
    \bibliography{main}
}

\clearpage
\setcounter{page}{1}
\maketitlesupplementary

\appendix

\begin{figure*}[t!]
    \centering
    \includegraphics[width=1\linewidth]{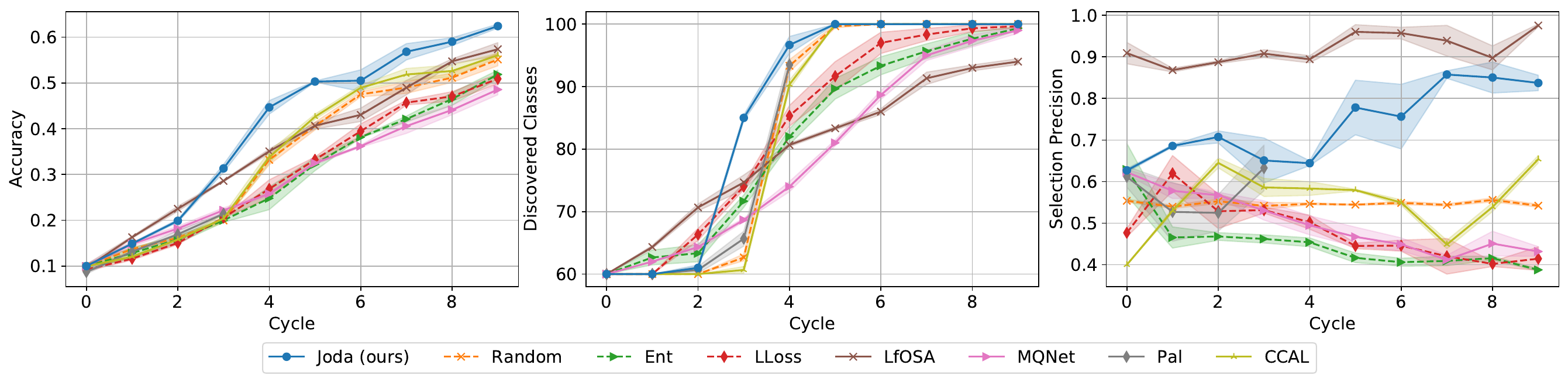}
    \caption{Experiments on ResNet50 and indicated standard errors on CIFAR-100 with Places365.}
    \label{fig:Restnet50}
\end{figure*}

\begin{figure*}[t]
    \centering
    \includegraphics[width=1\linewidth]{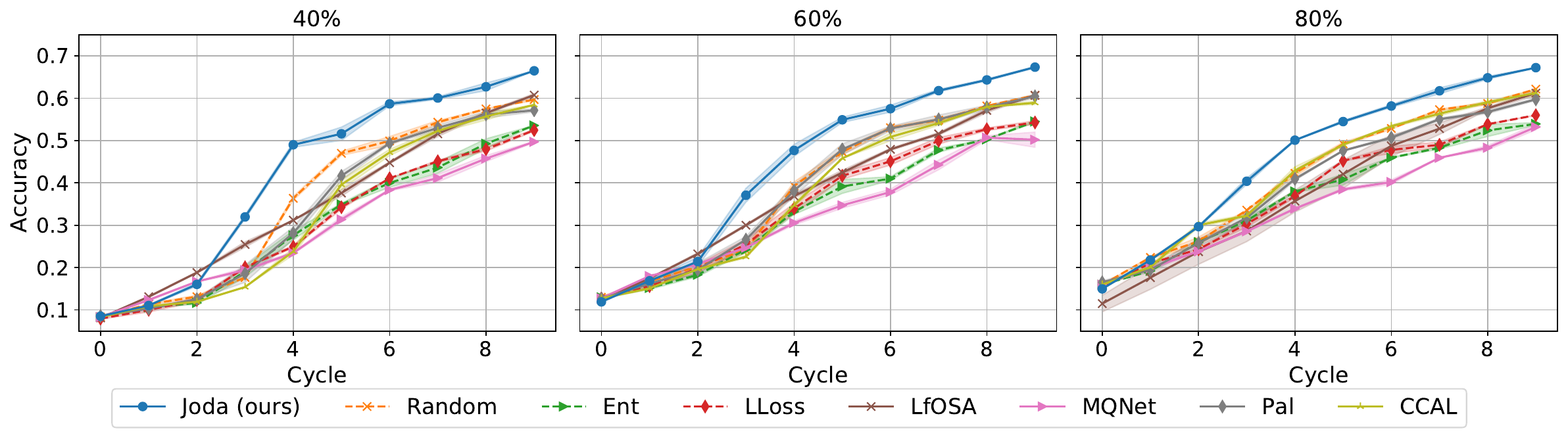}
    \caption{Different InD to discovery set splits  with ResNet18 and indicated standard errors on CIFAR-100 with Places365 - InD percentage left to right: 40\%, 60\% and 80\%.}
    \label{fig:CompInDRate}
\end{figure*}

\begin{figure*}[t!]
    \centering
    \includegraphics[width=1\linewidth]{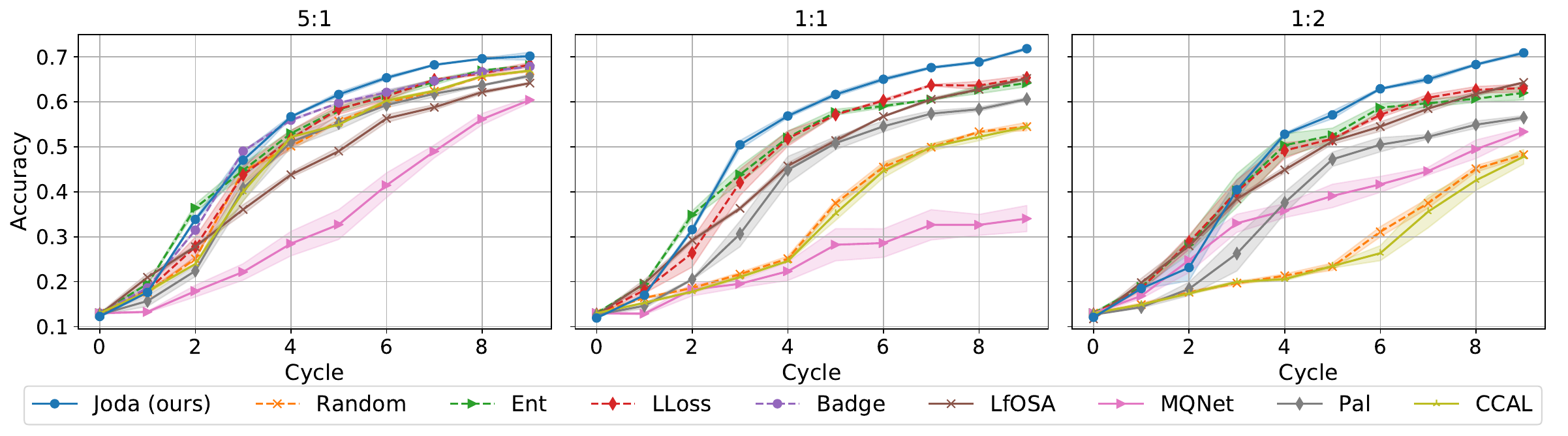}
    \caption{Different InD and OOD ratios with ResNet18 and indicated standard errors on CIFAR-100 with MNIST - InD to OOD ratio left to right: 5:1, 1:1, 1:2. }
    \label{fig:CompOODRate}
\end{figure*}

\section{Experimental Details}
\label{app:experimentdDetails}

In this section, we provide an overview of the parameters and implementation
details for the methods and experiments conducted.
For the full scenario experiments involving \cref{fig:Cifar10,fig:Cifar100,fig:ImageNet}, we initially designated 60\% of the dataset classes as InD data, while the remaining classes were considered discoverable. The threshold $t_e$ for adding a novel class from the discovery set $D$ to the InD data $I$ is derived from a uniformly distribution query size. For CIFAR-10 \cite{cifar}, it is set to 100, while for CIFAR-100 \cite{cifar} and TinyImageNet \cite{le2015tiny}, it is set to 50 and 25, which is slightly higher than the uniformly distributed query size. The classes are enumerated consecutively to the percentage. The same procedure is utilized for respective percentages in \cref{fig:CompInDRate}. All experiments were conducted with three different seeds. Only the scenario studies and the TinyImageNet experiments were conducted using two seeds. 
Across all datasets, we employed random horizontal flipping and random cropping with a padding size of 4, following common practices in \cite{Yoo2019,Kim2021,Caramalau2021}. Following these works, the initial labeled pool sizes for CIFAR-10 and CIFAR-100 were set to 1000 and 2000, respectively, before filtering out the discoverable classes. For TinyImageNet \cite{le2015tiny}, the size is established at 40 samples per class according to \cite{Ning2022}. The query sizes were set to 1000 for CIFAR-10 and 2500 for CIFAR-100 and TinyImageNet. 
In the case of CIFAR-10 and CIFAR-100, we utilized the same augmentations and initial pool size as \citet{Yoo2019}, specifically employing horizontal flipping and random cropping with a padding of 4. Similar augmentations were applied for TinyImageNet, as they only marginally differed from those examined by \citet{le2015tiny}. We employed a ResNet18 \cite{He2016} as the task model, along with the parameters and model modifications advised for the benchmark datasets \cite{cifar10Repo,cifar100Repo}. These modifications have also been utilized in previous AL works \cite{Yoo2019, Kim2021,Caramalau2021}. TinyImageNet, a subset of ImageNet with an image size of (64, 64) and only 200 classes compared to the original 1000 classes, was subject to the same optimizer and scheduler settings as those chosen for CIFAR-10. We trained the task model from scratch for 200 epochs in each cycle and proceeded with the model, achieving the best performance on the validation dataset for evaluation and sample selection. 

The method implementations are based on the official repositories of the respective models: LfOSA \cite{Ning2022}, CCAL \cite{Du2021}, Pal \cite{Yang2023}, and MQNet \cite{Park2022}. We made modifications to the dataloader and dataset definition for the contrastive training in Pal, CCAL, and MQNet, considering only far-OOD data as OOD data for the method while ignoring discoverable classes. We tried to adhere to the parameters suggested by the authors as much as possible. However, we decided to reduce the number of epochs for CCAL and MQNet to 100 due to the increased dataset size, resulting in increased iterations per epoch. For Badge \cite{Ash2020}, we also followed the official implementation. Regarding SISOMe \cite{Schmidt2024} applied for the selection in \ours, we used the suggested parameters for the sigmoids of 100, 1000, 0.001, and 0.001 and maintained these settings for all experiments. Additionally, we set $\lambda_{\text{OE}}=0.5$ for all experiments. 

The methods EOAL \cite{Safaei2024} and AGD \cite{Ma2024} cannot be applied to the OSDAL scenario. EOAL required one additional model per class, which is hard to transfer to unknown classes. AGD focuses on selecting only novel classes and does not select already known classes, which is hard to adapt to select known and novel classes. A naive adaption did not show meaningful results.  

Regarding the OOD dataset, we utilized the "randn" function from PyTorch \cite{pytorch} to generate random noise data. The amount of noise OOD data $O$ is chosen to be equal to the amount of InD data $I$. Additionally, we provided three different sizes of MNIST as OOD dataset: the validation set comprising 10000 images, the complete training set comprising 60000 images, and lastly, we have augmented the training set by rotating the first 40000 images of MNIST, resulting in a dataset of 100000 images. These MNIST datasets stem the OOD data for \cref{fig:CompOODRate}. Further, the MNIST validation data used for \cref{fig:Cifar10,fig:Cifar100} as suggested by \cite{yang2022openood}. For the much larger TinyImageNet dataset in \cref{fig:ImageNet} the MNIST training set was used. For CIFAR-10 and CIFAR-100 in \cref{fig:Cifar10,fig:Cifar100}, the MNIST validation set was utilized as OOD data. For ImageNetC-800 \cite{Hendrycks2019}, we used the 800 ImageNetC \cite{Hendrycks2019} classes, which are not present in TinyImageNet with the motion blur 2 perturbation. All experiments have been conducted on Nvidia V100 GPU with 32 GB RAM and 8 CPU Cores.

\section{Additional Model Evaluation}
\label{ap:AdditionalModelEvaluation}
To validate the usability of \ours{} for different kinds of models, we conducted additional experiments with a ResNet50 for CIFAR-100 with Places365. We present the results in \cref{fig:Restnet50}. For the larger model, all methods show a related performance compared to the ResNet18 model in \cref{fig:Cifar100}. However, CCAL and LfOSA show less overlap in accuracy. \ours{} remains in the first place for accuracy and class discovery.

\begin{figure*}[t!]
    \centering
    \includegraphics[width=1\linewidth]{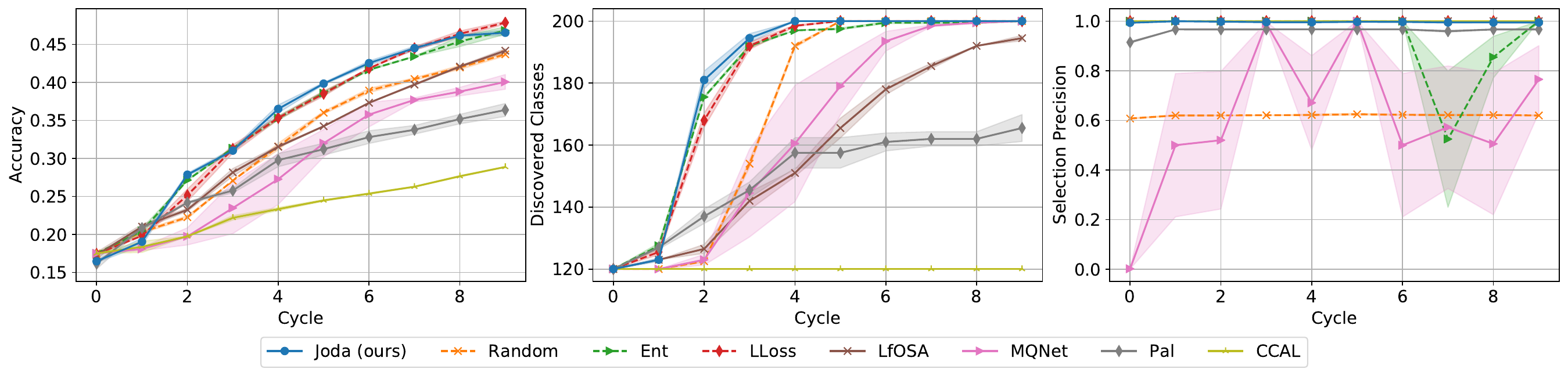}
    \caption{Experiments on ResNet18 and indicated standard errors on TinyImageNet with Random Noise.}
    \label{fig:ImageNetP}
\end{figure*}

\section{Scenario Variation}
\label{ap:ScenarioVariation}
To evaluate the effect of different data split ratios, we conduct experiments on the InD to discovery ratios, as well as InD to OOD ratios. 
\subsection{Varying Discoverable Set Size}
Firstly, we analyze varying amounts of initially known categories in \cref{fig:CompInDRate}. While for $40\%$, the open-set AL methods maintain a gap to the classic AL methods, the gap decreases when more data is added. Especially, LfOSA reports a drop for $80\%$ InD classes. \ours{} maintains the highest performance overall amounts. 
\subsection{Varying OOD Set Size}
Subsequently, we are examining the impact of various OOD dataset sizes. Our experiments with different OOD datasets indicated an influence of the OOD dataset size on the results, particularly for the random selection profits from low OOD amounts for selection precision. In \cref{fig:CompOODRate}, we compare different OOD to InD ratios. As expected, the classic AL approach suffers performance drops with rising OOD amounts, while the open-set AL methods, in particular LfOSA, maintain their performance. \ours{} achieve the highest accuracy over all ratios and extends the margin for increased OOD data. The result underlines the robust behavior of \ours{} for various dataset splits. %

\begin{figure}[h!]
    \centering
    \includegraphics[width=0.8\linewidth]{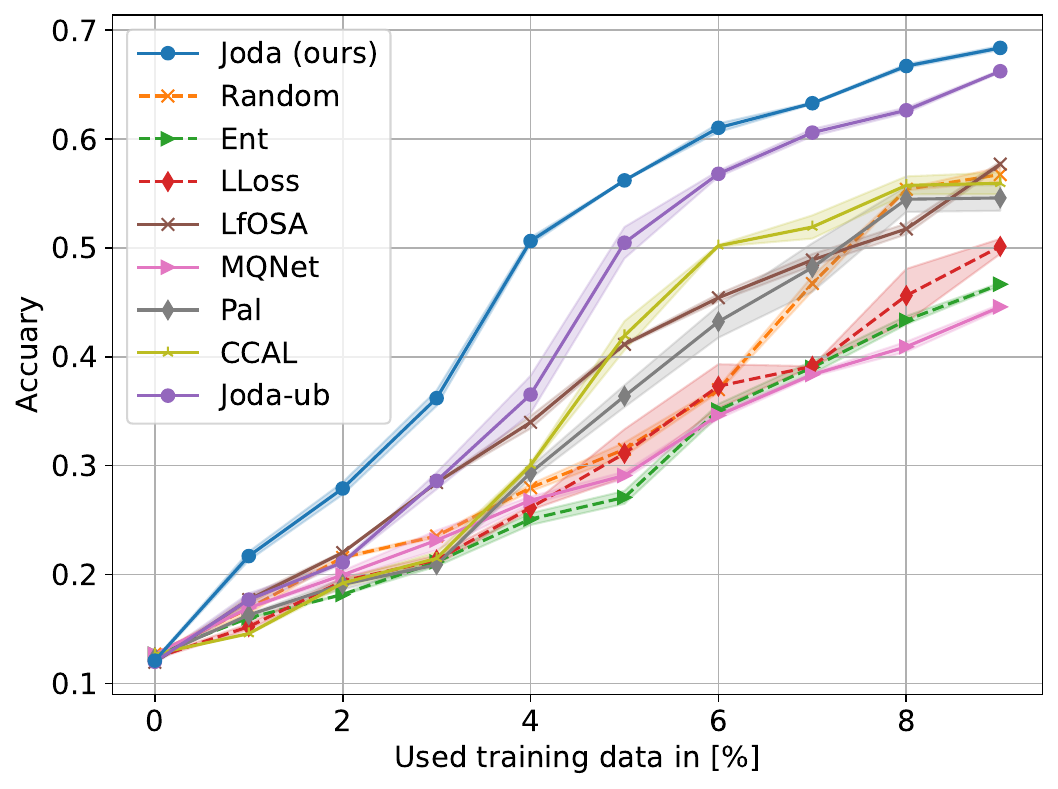}
    \caption{Comparison of an $U_I$ to $U_D$ ratio of 2:1 for CIFAR-100 and Places365 with indicated standard errors using ResNet18.}
    \label{fig:UnbalancedCounts}
\end{figure}

\begin{figure*}[t!]
    \centering   \includegraphics[width=1\linewidth]{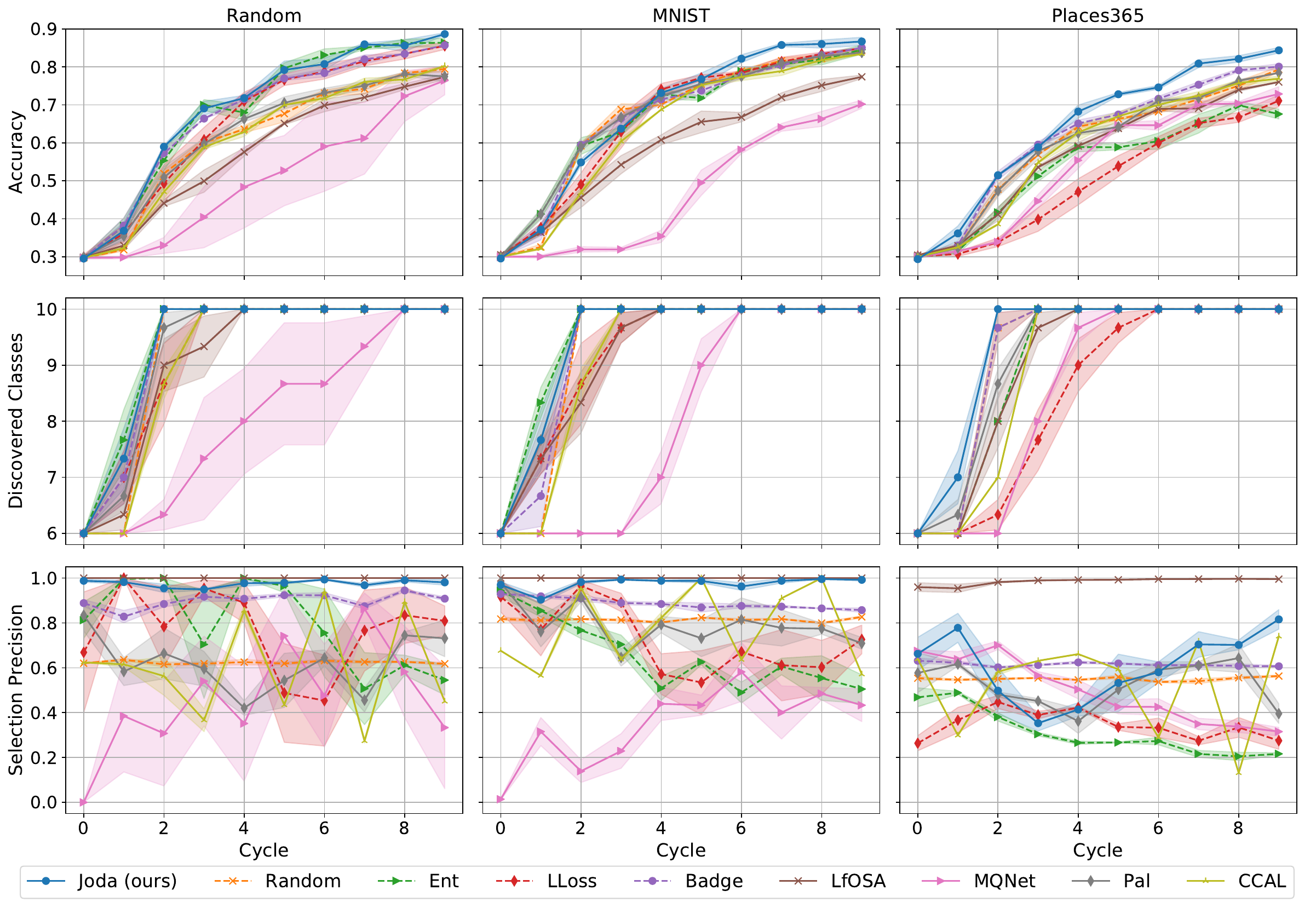}
    \vskip -0.3cm
    \caption{Comparison for CIFAR-10 with ResNet18 and indicated standard errors. From top to bottom: Accuracy, Class Detection, and Selection Precision. OOD datasets from left to right: Random, MNIST, and Places365.}
    \label{fig:Cifar10}
    \vskip -0.2cm
\end{figure*}

\subsection{Unbalanced Class Samples}

In the most experimental setting, the number of samples of already known and discoverable classes are roughly equal. However, in some scenarios, the number of samples of discoverable classes might be lower as they have not been revealed yet. In \cref{fig:UnbalancedCounts}, we compare the different methods a setting where the discoverable classes are less frequent. It can be seen that \ours{} shows a strong resistance to the unbalanced classes and increases the gap to the setting of Places365 shown in \cref{fig:Cifar100}. In addition, to our \ours{} we show, the effect of our novel introduced balancing in \cref{eq:balanceFactor}. Joda-ub presents the unbalanced version, without this factor. While Joda-ub manages to outperforms other AL methods, there remains still a gap to \ours{}.

\begin{figure*}
    \centering
    \includegraphics[width=\linewidth]{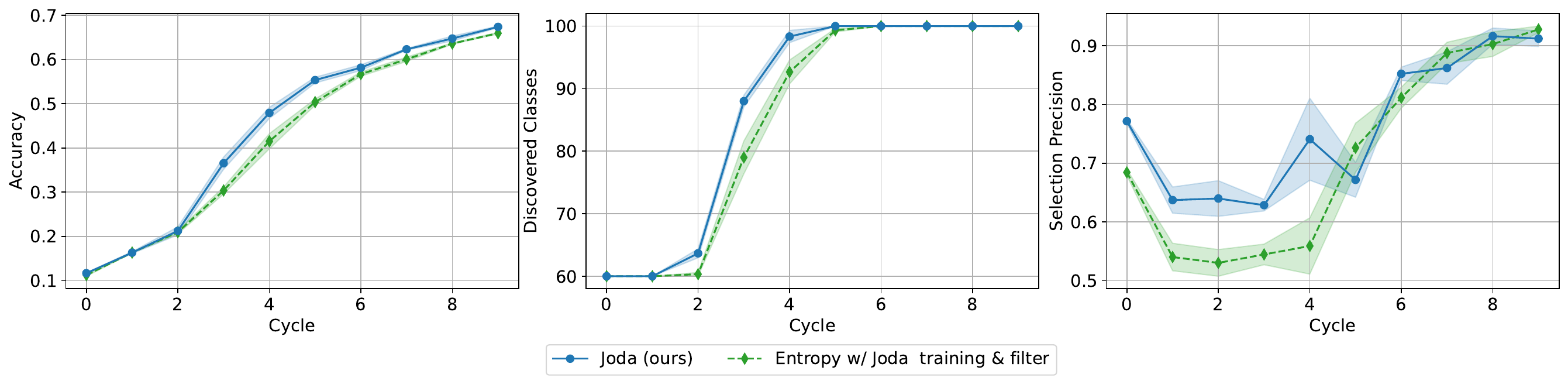}
    \caption{Comparison of different query sizes for CIFAR-100 and Places365 with indicated standard errors using ResNet18.}
    \label{fig:EntropyVsJoda}
\end{figure*}

\section{Further TinyImageNet Experiments}
\label{ap:OverlapScenario}
In addition to the experiments conducted in \cref{sec:Experiments}, where we evaluate TinyImageNet with MNIST, ImageNetC-800, and Places365, we examine the behavior for random noise. In \cref{fig:ImageNetP}, we present these results with accuracy, class discovery, and selection precision. Compared to CIFAR-10 and CIFAR-100 presented in \cref{fig:Cifar10} and \cref{fig:Cifar100}, AL approaches and \ours{} perform much better in terms of accuracy as well as selection precision. Given the larger and more realistic samples, random noise behaves much more abnormal, which can be easier detected by metrics instead of learned latent spaces. In this experiments \ours{} achieved a the highest class discovery with a perfect selection precision of 1. In cycles 2-8 \ours{} reports the highest accuracy and is only slightly eclipsed in the last cycle and in the starting cycles where the data is not explored yet. In general, \ours{} achieves top performance for all three metrics in almost all measurement points.

\section{Effect of Selection Size}
\label{ap:SelSize}

\begin{figure*}
    \centering
    \includegraphics[width=1\linewidth]{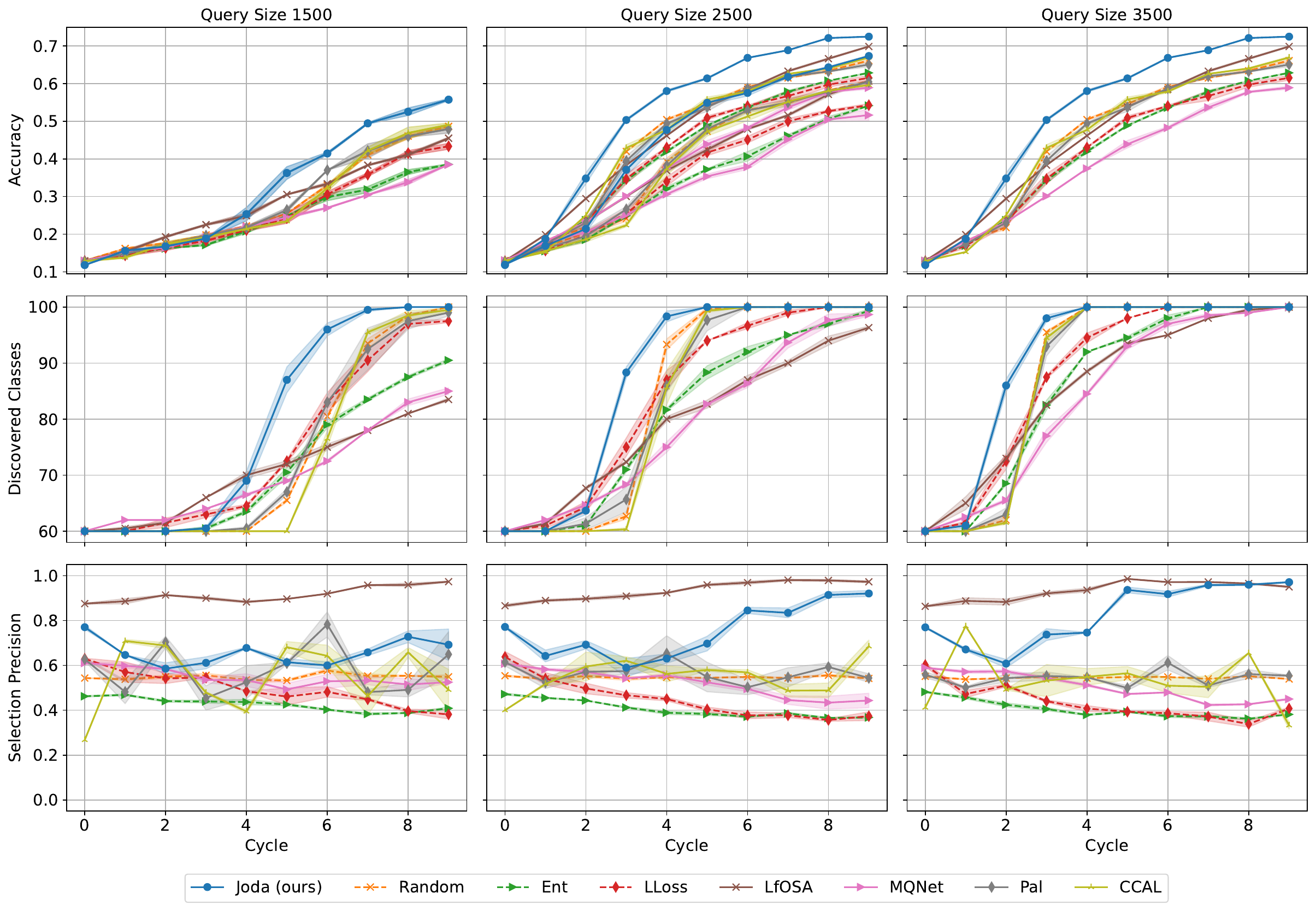}
    \caption{Comparison of different query sizes for CIFAR-100 and Places365 with indicated standard errors using ResNet18.}
    \label{fig:CompSelSize}
\end{figure*}

An essential aspect of AL is the number of selected samples per cycle. The commonly used setting selects 2500 samples for CIFAR-100 as highlighted in \cref{app:experimentdDetails}. While most work in open-set AL do not investigate, this influence of varying selection sizes we investigate the effects of a lower and higher selection size in \cref{fig:CompSelSize}. In can be seen, that \ours{} shows in general a high performance over all selection sizes and maintains the difference to other methods. Based on the entanglement of training and filtering, higher selection ratios lead a faster increase of selection precision.

\section{Ablation Study on Selection Method}
While Joda entangles the training, filtering and selection in an enhanced manner to avoid additional data access and axillary model or ensemble requirement, its structure remains flexible. Based by this design \ours{} remains open to methods improvements. In \cref{fig:EntropyVsJoda}, we replace the selection of \ours{} by and MC Dropout Entropy selection. While it can be seen, that \ours{} outperforms the Entropy selection, it should be notes, that \ours's training and filtering strategy massively improves the performance of the Entropy selection compared showcasing the effectiveness of ours entangles setup.

\section{Experiments on CIFAR-10}
\label{ap:cifar10}
In \cref{fig:Cifar10}, we present our results for CIFAR-10. The columns represent the OOD datasets from left to right: random noise (Random), MNIST, and Places365. The visualizations depict the accuracy, class discovery, and selection precision from top to bottom.
The classic AL methods, Ent \cite{Gal2015} and LLoss \cite{Yoo2019}, demonstrate decent performance for random noise as OOD but decrease in performance with the complexity of the OOD data and get outperformed by others for Places365. Badge \cite{Ash2020}, which incorporates diversity, achieves decent overall performance in OOD datasets. Additionally, classic AL methods produce inconsistent results for precision selection. However, together with Random selection, they benefit from a small OOD ratio, as in the case of MNIST.
Notably, the open-set methods based on contrastive learning CCAL \cite{Du2021} and MQNet \cite{Park2022} achieve inconsistent results. CCAL performs well for MNIST and Places365, while MQNet has difficulties with far-OOD Random and MNIST. For Places365, MQNet achieves good performance and class separation but suffers in selection precision.
LfOSA \cite{Ning2022}, utilizing an additional classifier, demonstrates good precision but struggles with class discovery.
However, Pal \cite{Yang2023}, CCAL, and MQNet improve their selection precision but do not surpass LfOSA or \ours{} in this aspect.
\ours{} exhibits a strong performance and early class discovery with stable selection precision. For Places365, as OOD data \ours{} initially falls behind regarding selection precision but recovers after a few cycles. In comparison to CIFAR-100, the gap between the methods is smaller, given the reduced number of classes. Overall, \ours{} showed the best combination of accuracy, class discovery, and selection precision.

\end{document}